\RequirePackage[hyphens]{url}
\documentclass[10pt,twocolumn,letterpaper]{article}

\usepackage{graphicx}
\usepackage{subcaption}
\usepackage{float}
\usepackage[justification=raggedright]{caption}	
\usepackage{lscape}                                         
\usepackage{wrapfig}

\usepackage[lined,ruled,linesnumbered]{algorithm2e}
\usepackage{animate} 

\usepackage{booktabs}                   
\usepackage{multirow}

\usepackage{makecell}

\usepackage{paralist}
\usepackage{enumitem}
\setlist[enumerate]{itemsep=0mm}

\usepackage{bm}                          
\usepackage{epsfig}                      
\usepackage{graphicx}                  
\usepackage{times}
\usepackage{mathptmx}
\usepackage{mathtools}
\usepackage{amssymb,amsmath}   

\usepackage{units}
\usepackage{color}

\usepackage[pagebackref,breaklinks,colorlinks,linkcolor=blue,citecolor=blue,bookmarks=false]{hyperref}

\usepackage{xspace}
\usepackage[table]{xcolor}
\usepackage{setspace}
\usepackage{grfext}
\PrependGraphicsExtensions*{.jpg,.png,.PNG}

\usepackage{gensymb}

\usepackage{soul}
\usepackage{svg}

\usepackage[accsupp]{axessibility} 

\usepackage[final]{changes}

\DeclareMathAlphabet{\altmathcal}{OMS}{cmsy}{m}{n}
\DeclareMathAlphabet{\mathbfit}{OT1}{ptm}{bx}{it}

\newlength\paramargin
\newlength\figmargin

\newlength\secmargin
\newlength\figcapmargin
\newlength\tabcapmargin

\newcommand{\mpage}[2]
{
\begin{minipage}{#1\linewidth}\centering
#2
\end{minipage}
}

\newcommand{\topic}[1]
{
\vspace{1mm}\noindent\textbf{#1}
}

\long\def\ignorethis#1{}

\newbox\jsavebox%

\makeatletter
\newcommand{\providelength}[1]{%
  \@ifundefined{\expandafter\@gobble\string#1}
   {
    \typeout{\string\providelength: making new length \string#1}%
    \newlength{#1}%
   }
   {
    \sdaau@checkforlength{#1}%
   }%
}

\newcommand{\sdaau@checkforlength}[1]{%
  \edef\sdaau@temp{\expandafter\sdaau@getfive\meaning#1TTTTT$}%
  \ifx\sdaau@temp\sdaau@skipstring
    \typeout{\string\providelength: \string#1 already a length}%
  \else
    \@latex@error
      {\string#1 illegal in \string\providelength}
      {\string#1 is defined, but not with \string\newlength}%
  \fi
}
\def\sdaau@getfive#1#2#3#4#5#6${#1#2#3#4#5}
\edef\sdaau@skipstring{\string\skip}
\makeatother

\usepackage[capitalize]{cleveref}
\crefname{section}{Sec.}{Secs.}
\Crefname{section}{Section}{Sections}
\Crefname{table}{Table}{Tables}
\crefname{table}{Tab.}{Tabs.}

\def\xi{\mathbf{x}_i}

\graphicspath{{[CVPR 2023] Template/figures}, {examples}}

\makeatletter
 
\def\@fnsymbol#1{\ensuremath{\ifcase#1\or \dagger\or \ddagger\or
\mathsection\or \mathparagraph\or \|\or **\or \dagger\dagger
\or \ddagger\ddagger \else\@ctrerr\fi}}
\makeatother

\graphicspath{{figures/}}

\usepackage[pagenumbers]{cvpr} 
\usepackage{amsmath}
\usepackage[accsupp]{axessibility}
\usepackage[vskip=1em,font=itshape,leftmargin=2em,rightmargin=2em]{quoting}
\usepackage{lipsum}

\newcommand\blfootnote[1]{%
  \begingroup
  \renewcommand\thefootnote{}\footnote{#1}%
  \addtocounter{footnote}{-1}%
  \endgroup
}

\def\confName{----}
\def\confYear{----}

\begin{document}

\title{
Seeing the World through Your Eyes
}

\author{Hadi Alzayer*~~~  Kevin Zhang* ~~~ Brandon Feng ~~~
Christopher Metzler~~~  Jia-Bin Huang \\
University of Maryland, College Park\\
\url{https://world-from-eyes.github.io/}
}

\twocolumn[{
\renewcommand\twocolumn[1][]{#1}
\maketitle
\begin{center}
\centering

\includegraphics[width=\linewidth]{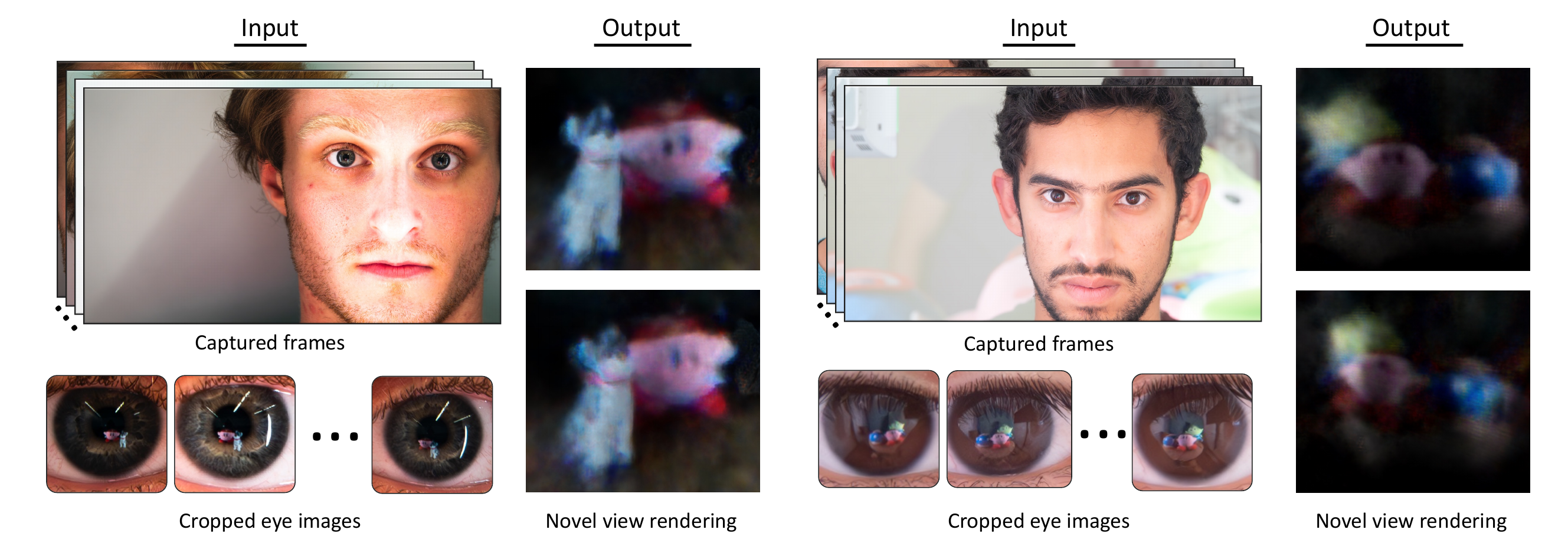}





\captionof{figure}{
\textbf{Radiance field reconstruction using eye reflections.} 
The human eye is highly reflective. 
We show that from a sequence of frames that capture a moving head, we can reconstruct and render the 3D scene of what the person is observing using only the reflections off their eyes.
}

\label{fig:teaser}
\end{center}

}]

\maketitle
\begin{abstract}

The reflective nature of the human eye is an underappreciated source of information about what the world around us looks like. 
By imaging the eyes of a moving person, we can collect multiple views of a scene outside the camera's direct line of sight through the reflections in the eyes. 
In this paper, we reconstruct a 3D scene beyond the camera's line of sight using portrait images containing eye reflections.
This task is challenging due to 
1) the difficulty of accurately estimating eye poses and 
2) the entangled appearance of the eye iris and the scene reflections.
Our method jointly refines the cornea poses, the radiance field depicting the scene, and the observer's eye iris texture. 
We further propose a simple regularization prior on the iris texture pattern to improve reconstruction quality. 
Through various experiments on synthetic and real-world captures featuring people with varied eye colors, we demonstrate the feasibility of our approach to recover 3D scenes using eye reflections.
\blfootnote{*Equal contribution}
\end{abstract}


\section{Introduction}
\label{sec:introduction}
\begin{quoting}
The only true voyage of discovery ... would be not to visit strange lands but to possess other eyes, to behold the universe through the eyes of another ... -- Marcel Proust, 1927
\end{quoting}
The human eye is a remarkable organ that enables vision and holds valuable information about the surrounding world.
While we typically use our own eyes as two {\it lenses} to focus light onto the photosensitive cells composing our retina, we would also capture the light reflected from the cornea if we look at someone else's eyes.
When we use a camera to image the eyes of another, we effectively turn their eyes as a set of {\it mirrors} in the overall imaging system.
Since the light that reflects off the observer's eyes share the same source as the light that reaches their retina, our camera should form images containing information about the world the observer sees.

Prior studies have explored recovering a panoramic image of the world the observer sees from an image of two eyes~\cite{world_in_eye_cvpr, world_in_eye_journal}.
Follow-up works have further explored applications such as personal identification~\cite{Nishino2005UsingER,bystander_id}, detecting grasp posture~\cite{zhang2022reflectouch}, focused object estimation~\cite{takemura2014estimation}, and relighting~\cite{Nishino2004EyesFR}.
Given the recent advancements in 3D vision and graphics, we wonder: 
Can we do more than reconstruct a single panoramic environment map or recognize patterns?
Is it possible to recover the world seen by the observer in full 3D? 

In this paper, we answer these questions by reconstructing a 3D scene from a sequence of eye images.
We start from the insight that our eyes capture/reflect multi-view information as we naturally move our heads.
We draw inspiration from the classical imaging formulation proposed by~\cite{world_in_eye_cvpr} and integrate it with the recent advances in 3D reconstruction spearheaded by Neural Radiance Fields (NeRF)~\cite{mildenhall2020nerf}.
Unlike the standard NeRF capture setup, which requires a \emph{moving camera} to capture multi-view information (often followed by camera pose estimation), our approach employs a \emph{stationary camera} and extracts the multi-view cues from eye images under head movement.

While conceptually straightforward, reconstructing a 3D NeRF from eye images is extremely challenging in practice. 
The first challenge is source separation.
We need to separate the reflections from the intricate iris textures of human eyes. 
These complex patterns add a level of ambiguity to the 3D reconstruction process. 
Unlike the clear images of the scene typically assumed in standard captures, the eye images we obtain are inherently blended with iris textures. 
This composition disrupts the pixel correspondence and complicates the reconstruction process.
The second challenge is cornea pose estimation.
Eyes are small and hard to localize accurately from image observations.
The multi-view reconstruction, however, depends on the accuracy of their locations and 3D orientations.

To address these challenges, in this work, we repurpose NeRF for training on eye images by introducing two crucial components: 
a) texture decomposition, which leverages a simple radial prior to facilitate separating the iris texture from the overall radiance field, and 
b) eye pose refinement, which enhances the accuracy of pose estimation despite the challenges presented by the small size of eyes.

To evaluate the performance and effectiveness of our approach, we generate a synthetic dataset of a complex indoor environment with images that capture the reflection from a synthetic cornea with realistic texture. 
We further implement a real-world setup with multiple objects to capture eye images.
We conduct extensive experiments on synthetic and real-world captured eye images to validate several design choices in our approach.

Our primary contributions are as follows:
\begin{itemize}
\item {\bf New 3D reconstruction problem}. We present a novel method for reconstructing 3D scenes of the observer's world from eye images, integrating earlier foundational work with the latest advancements in neural rendering.
\item {\bf Radial prior for irises}. We introduce a radial prior for iris texture decomposition in eye images, significantly improving the quality of the reconstructed radiance field.
\item {\bf Cornea pose refinement}. We develop a cornea pose refinement procedure to alleviate the noisy pose estimates of eyes, which overcomes the unique challenge of extracting features from human eyes.
\end{itemize}
These advancements extend the current capabilities of 3D scene reconstruction through neural rendering to handle partially corrupted image observations obtained from eye reflections, opening up new possibilities for research and development in the broader area of accidental imaging~\cite{Bouman2017TurningCI, Torralba2012AccidentalPA, Sharma2021WhatYC, liu2023humans} to reveal and capture 3D scenes beyond the visible line-of-sight.

\begin{figure}
\begin{center}
\centering

\includegraphics[width=\linewidth, trim=0 0 0 0, clip]{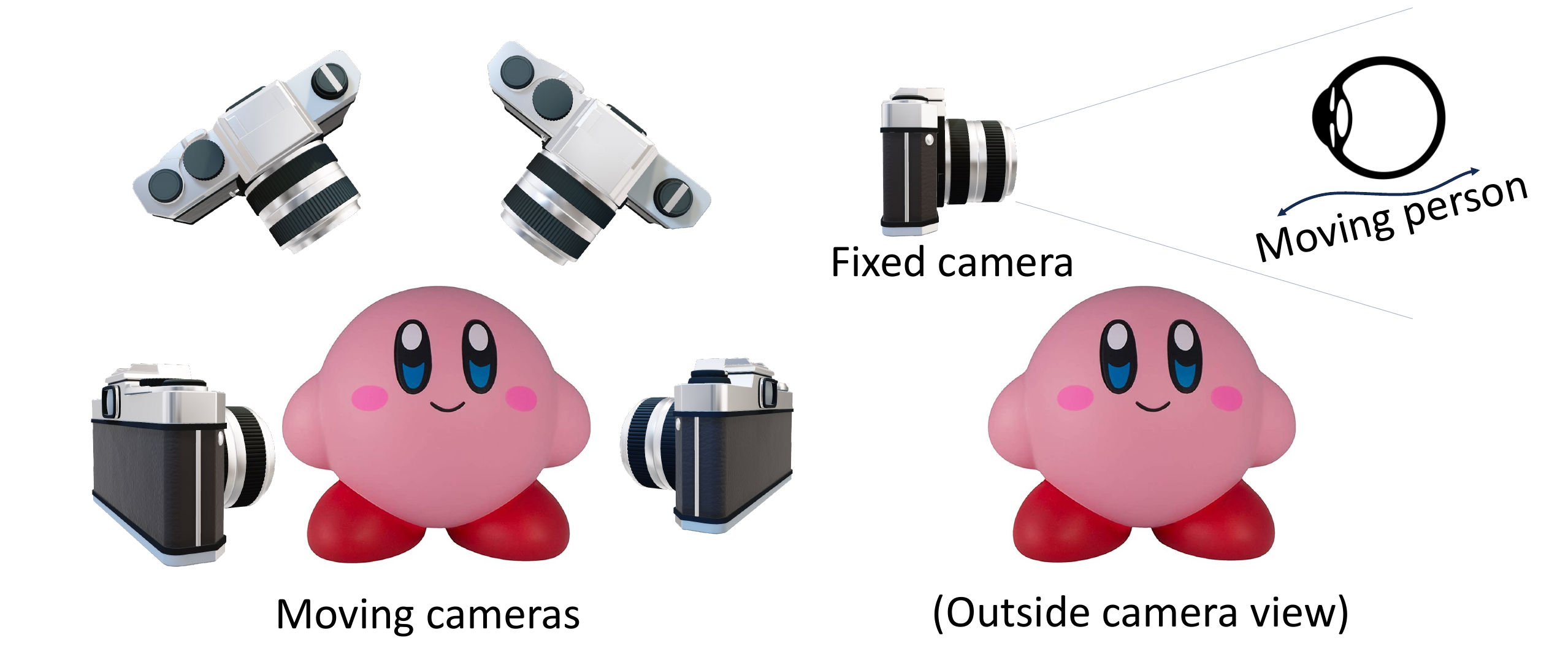}
\centering
\mpage{0.4}{\hspace{-20pt} 
(a) NeRF Setup} \mpage{0.2}{}
\mpage{0.3}{
(b) Our Setup}  \mpage{0.1}{}\hfill
\captionof{figure}{
\textbf{NeRF for non-line-of-sight scene.} 
The typical NeRF capture setup requires multiple posed images (e.g., captured from a moving camera) for reconstruction. 
In our setup, we gather multi-view information of the scene through light reflected from the eyes of a moving person.
}
\label{fig:why_figure}
\end{center}
\end{figure}

\section{Related Work}
\label{sec:related}



\topic{Catadioptric imaging.}
Catadioptric imaging uses a combination of lenses and mirrors for image capturing.
The word catadioptric is derived from {\it catoptrics} (related to the Greek words for specular and mirrors) and {\it dioptrics} (related to an Ancient Greek lens-like instrument).
In essence, catadioptric imaging seeks to leverage an additional (often curved) mirror to expand a lens-based imaging system's effective field of view.
Early studies in catadioptric imaging focused primarily on the design of the mirror profiles and their impact on the final image quality. 
\cite{Baker1998ATO} studied three design criteria of a catadioptric imaging system: the shape of the mirrors, the resolution of the cameras, and the focus settings of the cameras.
\cite{Swaminathan2003APO} provided a metric to quantify distortions and a method to minimize distortions in images acquired with a single viewpoint catadioptric camera. 
Moreover, a creative way to realize an accidental catadioptric imaging system is by treating human eyes as external curved mirrors~\cite{world_in_eye_journal}.
\cite{world_in_eye_cvpr} uses a single image of the eyes as a stereo system to identify pixel correspondences with epipolar geometry, even successfully identifying what the person is looking at. 
Another application of using human eyes as part of the imaging system is estimating light direction from the eyes to perform relighting~\cite{Nishino2004EyesFR, tsumura2003lightdir_from_eye}.
Our work draws inspiration from previous works on eye-based catadioptric imaging systems and further extends this concept to achieve 3D scene recovery through NeRF-based modeling. In particular, this paper introduces several new techniques to process catadiotrically captured eye images, such as learnable texture decomposition and refined iris estimations.

\topic{Neural radiance field.}
Neural radiance fields (NeRF)~\cite{mildenhall2020nerf} represent a significant milestone in novel view synthesis. 
NeRF adopts differentiable volume rendering to represent a 3D scene and uses neural networks to learn the density and color of each scene point. Following the success of NeRF, a plethora of follow-up works have been introduced to improve its rendering quality~\cite{barron2021mipnerf, barron2022mipnerf360}, ability to handle scene dynamics~\cite{park2021hypernerf,pumarola2020d,park2021nerfies,li2021neural,gao2021dynamic,liu2023robust}, inaccurate camera poses~\cite{lin2021barf, wang2021nerfmm, bian2022nopenerf, meuleman2023localrf,liu2023robust}, and rendering speed~\cite{yu2021plenoxels,attal2022learning,mueller2022instant}.
Our work uses NeRF to parametrize the unknown scene we wish to recover from eye reflections. 
In particular, we modify the training framework from nerfstudio~\cite{nerfstudio} to implement the NeRF-based scene reconstruction. 
We note that our input images are captured at a fixed viewpoint, which differs from the typical NeRF setup, which requires multi-view input with additional requirements of camera pose optimization.

\topic{Reflection removal.}
Removing reflections from captured images is a longstanding computational photography problem. 
The related literature on this topic can be summarized into two main categories: {\it multi-frame} and {\it single-image}.
Multi-frame reflection removal methods~\cite{Sinha2012ImagebasedRF,Xue2015ACA,Gai2012BlindSO,liu2020learning,liu2021learning} often exploit the differences of motion patterns between the background and reflection layers and impose various image priors as regularization.
Single-image reflection removal methods tend to exploit visual cues available in a single image, such as depth-of-field~\cite{Li2014SingleIL, Wan2016DepthOF},  defocus-disparity~\cite{Punnappurath2019ReflectionRU}, or learned image features~\cite{Zhang2018SingleIR}.
More recently, NeRF has emerged as a new tool for reflection removal, specifically under the multi-frame setting. 
Various NeRF-based methods have studied how to accurately model and extract specular reflections from shiny or metallic objects~\cite{Verbin2021RefNeRFSV, Zhang2021NeRFactorNF, glossyobjects2022, Dave2022PANDORAPN}.
Nerfren~\cite{Guo_2022_CVPR} demonstrates that by fitting two NeRFs to model the reflection and diffuse components of the scene separately, reflections from planar surfaces like mirrors can be removed and re-rendered as a separate 3D scene.
Due to the simplicity of planar reflections, Nerfren achieves the joint learning of reflection and diffuse components by simply aggregating predictions from two NeRF models together (reflection and diffuse) weighted by alpha-compositing.
In this work, unlike prior works that focus on planar surface geometry, our object of interest (the human eye) has an inherently more complicated curved geometry, which necessitates us developing several modifications to the standard NeRF rendering workflow, which we will detail in the following sections.

\begin{figure}
\begin{center}
\centering

\includegraphics[width=0.5\linewidth, trim=0 0 0 0, clip]{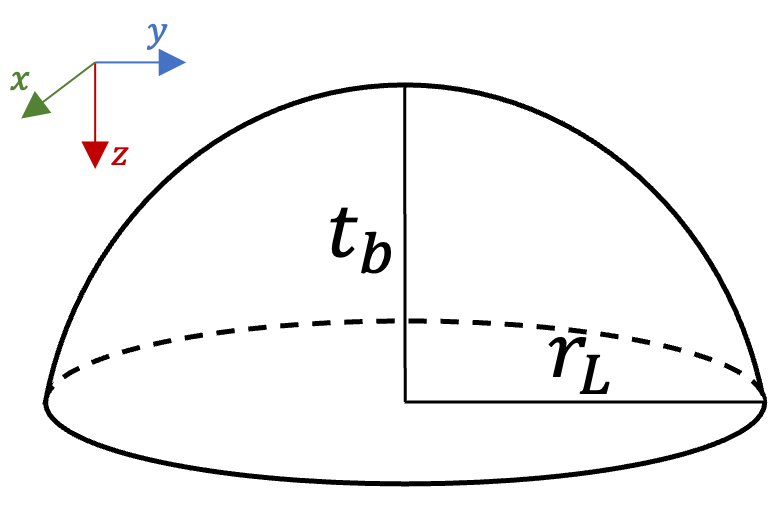}
\captionof{figure}{
\textbf{Cornea geometry.} 
The cornea can be modeled as an ellipsoid. 
The key fact that we exploit is that the cornea shape and size are largely consistent among adults, with similar eccentricity and curvature.
}
\label{fig:cornea}
\end{center}
\end{figure}

\begin{figure*}
\begin{center}
\centering

\includegraphics[width=\linewidth]{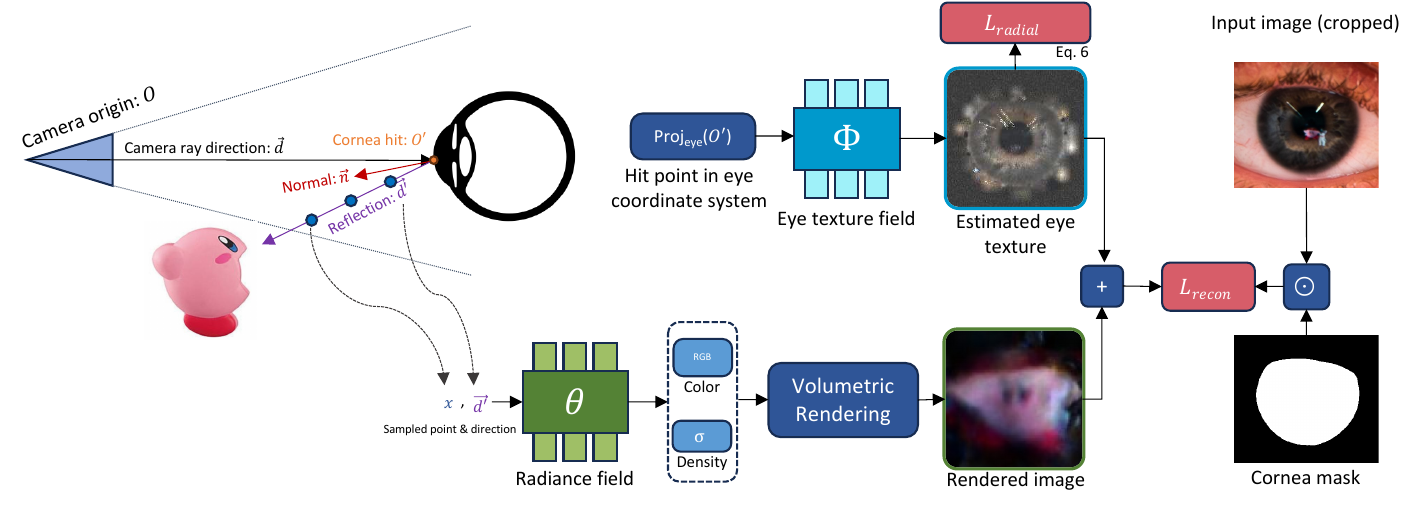}

\captionof{figure}{
\textbf{Joint optimization of radiance field and iris texture.} 
Standard NeRF rendering uses rays starting from the camera origin $O$ along a viewing direction $d$. 
In contrast, in our setup, we need to use rays that \emph{bounce off} the cornea.
The reflected ray origin $O'$ is where the initial camera ray intersects with the cornea, and the new ray direction $d'$ is the reflection of $d$ across the cornea's normal $\overrightarrow{n}$. 
Consequently, the eye image we observe is a composition of the iris texture and the reflected scene.
The composition hinders standard NeRF training due to the highly-detailed iris texture.
To address this issue, alongside the radiance field $\theta$, we train an \emph{eye texture field} $\Phi$ whose input is the projection of $O'$ on the eye coordinate system in the given image (Eq. \ref{eq:proj}). 
The eye texture field is computed relative to the eye in the current image, while the radiance field takes 3D points in the world coordinates. 
The outputs from volumetric rendering with $\theta$ and texture estimation with $\Phi$ are composited together to reconstruct the cornea image.
We apply a reconstruction loss $L_{recon}$.
We further regularize the texture field $\Phi$ with a radial loss $L_{radial}$ that encourages the estimated texture to be radially constant, reducing the absorption of scene regions into the eye texture.
}

\label{fig:how_figure}
\end{center}
\end{figure*}

\topic{Non-line-of-sight imaging.}
Non-line-of-sight (NLOS) imaging attempts to recover images of objects that are not directly visible from the camera's position or are obstructed by an object in the line of sight. 
The principle behind NLOS imaging is that one can use light reflected off a visible relay surface to record information about an object outside of the line of sight. 
The NLOS literature largely falls under two categories: {\it active} and {\it passive}.
Active NLOS imaging techniques involve using controlled light sources, such as lasers, and often rely on time-of-flight measurements to reconstruct the hidden scene.
\cite{velten2012recovering} introduced an ultra-fast imaging system that records light in flight, allowing the reconstruction of non-line-of-sight objects. \cite{OToole2018ConfocalNI, lindell2019wave, Kadambi2016OccludedIW} later presented various methods to improve the resolution of active NLOS imaging systems.
NeRF has also been recently introduced to active NLOS imaging, enabling more accurate reconstructions and better handling of noise~\cite{shen2021non,fujimura2023nlos}.
Passive NLOS imaging, on the other hand, exploits natural or ambient light and does not require a controlled light source.
\cite{Torralba2012AccidentalPA} introduced the concept of accidental pinhole and pinspeck cameras, which involves using incidental or unintentional imaging elements in the environment to capture unique perspectives or resolve hidden scenes.
\cite{Bouman2017TurningCI, Sharma2021WhatYC} analyzed shadow patterns and showed that these patterns contain sufficient information to reconstruct the shape of the hidden scene.
\cite{liu2023humans} use reflections captured by a thermal camera to reconstruct the 3D body pose of non-line-of-sight humans.
\cite{glossyobjects2022} recently presented Orca, which uses reflections from a glossy object observed in multi-view images to train a 3D NeRF for the surrounding environment.
In this context, our paper can be regarded as a special case of passive NLOS scene reconstruction. 
We focus on a specific relay surface (the human eye) and introduce techniques tailored for better information extraction from eye reflections.
Unlike Orca, which relies on images captured with a moving camera while the ``mirror" object is fixed, our method works for a stationary camera and uses the natural movement of the human eye ``mirrors", which is visualized in Figure \ref{fig:why_figure}.

\section{Background: Eye Model}
\label{sec:background}
The geometry of the human eye has been extensively studied~\cite{pandolfi2006model}. 
The major components that are visible in the eye are: the sclera; which is the white region of the eye, and the cornea; which includes the iris and the pupil. The cornea is covered by a thin film of tear fluid, making it highly reflective. 
As noted by \cite{world_in_eye_cvpr}, since the cornea can act as a mirror, the combination of a camera and the cornea resembles a \emph{catadioptric system}.  
In our work, we follow the eye model adopted by \cite{world_in_eye_cvpr} for the geometry we assume for the eye. 

The eye is modeled as a section of an ellipsoid, as illustrated in Figure \ref{fig:cornea}, which can be described using the equation \begin{equation}\label{eq:ellipsoid}
    \left(1-e\right)z^2 - 2 Rz \text{ + } r^2 = 0
\end{equation}
where $e$ is the eccentricity, $R$ is the radius of the curvature at the apex, and $r^2 = x^2 \text{ + } y^2$. 
For an adult with healthy eyes, on average $e$ is about $0.5$ and $R$ is about $7.8$ mm, with very little variation across different people. 
The bounds of the ellipsoidal section are determined by the distance from the apex to the base, labeled $t_b$ in Figure \ref{fig:cornea}.
From $r_L$, the radius of the base of the ellipsoidal section, known to be approximately $5.5$ mm in people, we can calculate $t_b$ as about $2.18$ mm. To compute the normal at each point on the surface of the ellipsoid, we can take the gradient of Eq. \ref{eq:ellipsoid} and get 
\begin{equation}\label{eq:normal}
   \overrightarrow{n}\left(x,y,z\right) = \left<2x, 2y, 2\left(1-e\right)z - 2R\right> 
\end{equation}
To compute the depth of the cornea, we first assume a weak perspective projection model, which is valid because the diameter of the base is at most 11 mm and thus small compared to the depth. 
Next, notice that the projection of the cornea onto the image will be an ellipse. 
Let the major radius of the ellipse be $r_{img}$.
Then under the projection model, the average depth of the cornea can be computed as
\begin{equation}\label{eq:depth}
    \text{depth}_{\text{avg}} = r_L \frac{f}{r_{img}}.
\end{equation}


\section{Method}
\label{sec:method}

\topic{Radiance field from reflection.}
NeRF trains a parameterized radiance field through volumetric rendering. 
Each pixel color is computed by sampling the color and density along a ray using a parameterized MLP $\theta$. 
In NeRF, the ray associated with a pixel starts from the origin of that image's camera, denoted by $O$, and the direction, denoted by $\overrightarrow{d}$, is towards the projection of that pixel on the camera plane. 
By training the radiance field this way, we can recover a 3D reconstruction of the scene. 
However, in our setup, what we are interested in is to do a reconstruction of the scene reflected from the person's eyes.
In Figure \ref{fig:how_figure}, we illustrate how we use the rays reflected from the eye.
The reflected ray starts with the origin where the camera ray intersects with the cornea at $O'$, and in the direction of the reflected ray  $\overrightarrow{d'}$ instead of using $O$ and $\overrightarrow{d}$.
We compute the reflected ray explicitly using the standard reflection equation:
\begin{align}
    \overrightarrow{d'} = \overrightarrow{d} - 2\left(\overrightarrow{n} \cdot \overrightarrow{d} \right) \overrightarrow{n},
\end{align}
where $\overrightarrow{n}$ is the normal at the hit point $O'$. 
Note we only need to compute the hit points and normals once before training for pixels associated with the cornea. 
Since we model the cornea geometry as an ellipsoid, we directly compute the hit points and normals using closed-form ellipsoid ray intersection formulas during the data processing step.

\topic{Texture decomposition}
Since the target images are the scene reflections off the cornea, training NeRF naively cause the output radiance field of mixing scene geometry and iris texture. 
To recover only the scene geometry in the radiance field, we jointly optimize a 2D field $\Phi$ to learn the eye texture. 
We assume that the iris texture remains the same across the different views while the person moves, while the scene reflections vary.
For each pixel, the input to the 2D texture field is the pixel coordinate projected on the eye in the input image 
\begin{align}
\label{eq:proj}
    \text{proj}_{\text{eye}}\left(y, x\right) = \left(\frac{y - c_y}{r_{img}}, \frac{x - c_x}{r_{img}} \right)
\end{align}
where $\left(c_x, c_y\right)$ is the coordinate of the center of the cornea, and $r_{img}$ is the observed cornea radius. This parameterization enforces the texture field to naturally learn the invariant regions of the cornea, while the radiance field learns the 3D geometry of the scene. 

However, when a part of the scene does not display considerable motion across the training views, it can be ``absorbed" as part of the texture instead of the 3D scene. 
To resolve this issue, we propose a radial regularization that encourages radial symmetry of the recovered texture. 
We implement the loss by randomly sampling a rotation matrix $\Tilde{R}$, and penalize the model on the color deviation between coordinate $p$ and coordinate $\Tilde{R}p$ as follows:
\begin{align}
    L_{radial}\left( p \right) = \lambda_{radial} \|\Phi \left( p \right)- \Phi\left(\Tilde{R}p\right) \|_2^2 
\end{align}
where $\lambda_{radial}$ is the weight of the radial loss. 
While the iris is not perfectly radially constant, the simple radial loss effectively removes the scene reflection while maintaining an accurate estimated texture.

\begin{figure*}[tp]
\begin{center}
\centering

\mpage{0.95}{\includegraphics[width=\linewidth, trim=0 0 0 0, clip]{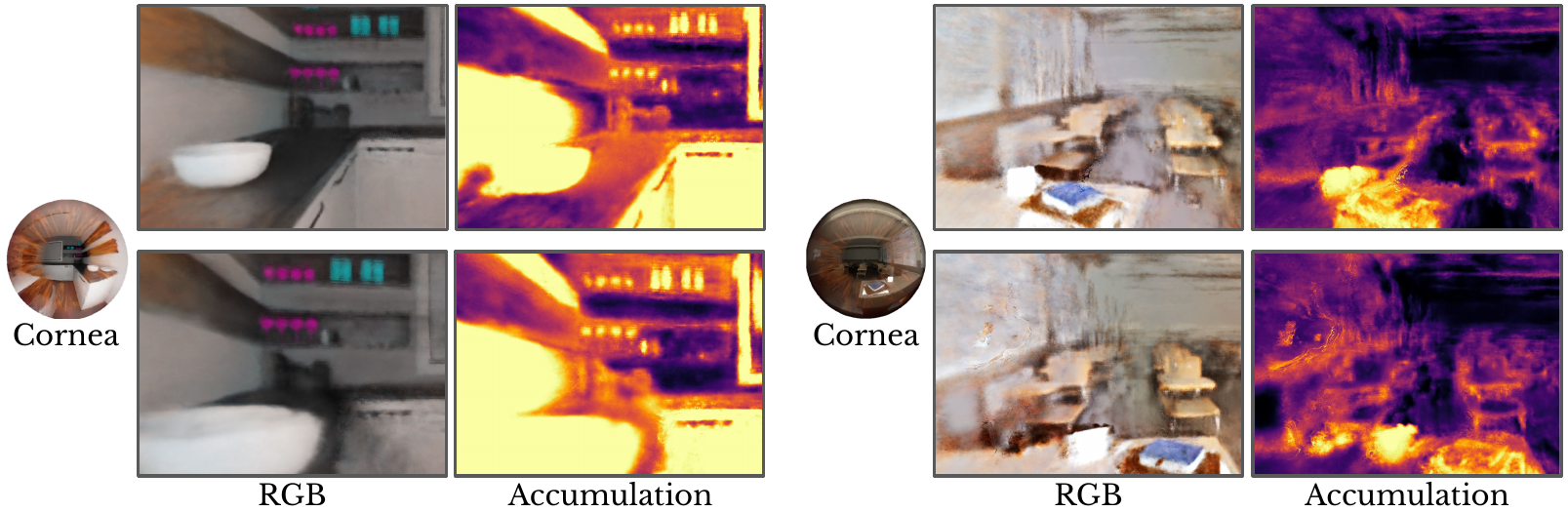}}
\mpage{0.95}{\includegraphics[width=\linewidth, trim=0 0 0 0, clip]{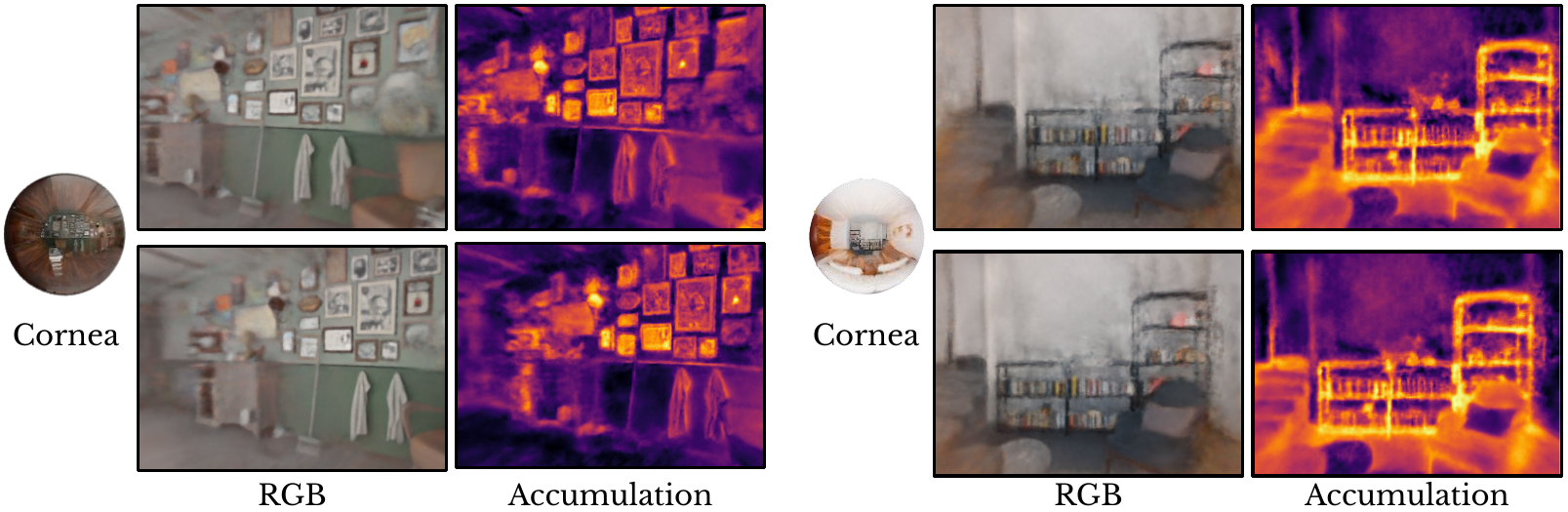}}
\mpage{0.95}{\includegraphics[width=\linewidth, trim=0 0 0 0, clip]{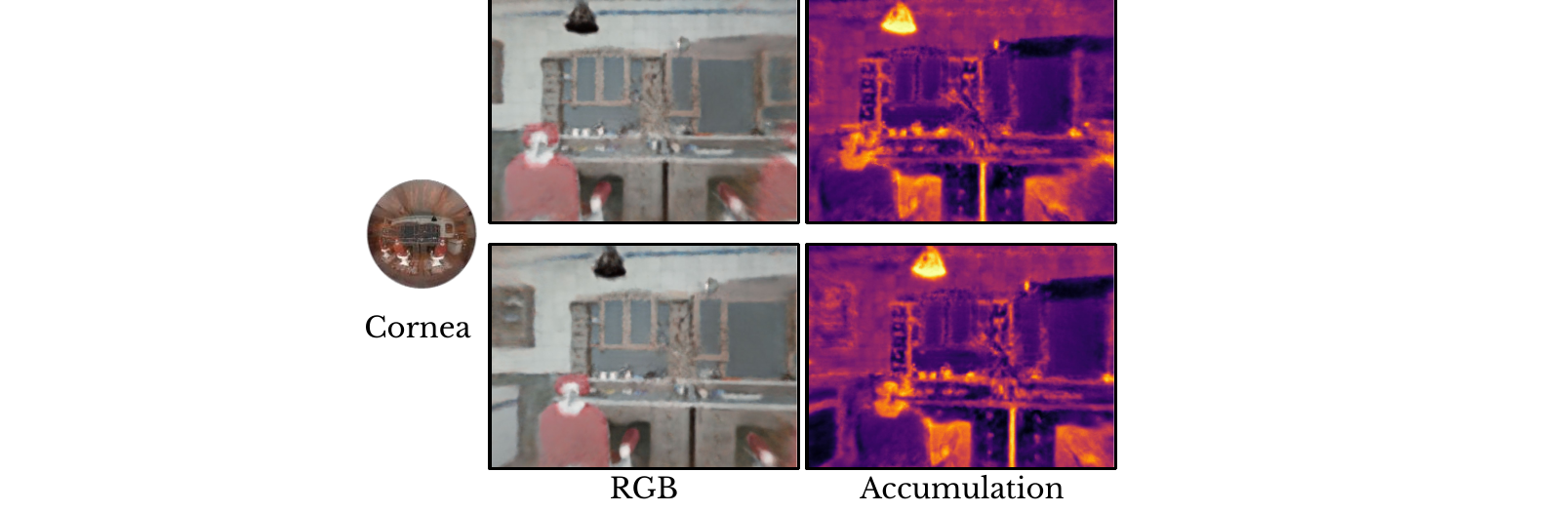}}
\captionof{figure}{
\textbf{Qualitative synthetic results.} 
We show that our method can achieve reasonable reconstructions from challenging measurements in simulation. 
We demonstrate that our method can reconstruct the 3D geometry of the scene by visualizing the accumulation of the learned radiance fields with respect to the camera poses.
The accumulation is defined as the integral of the density along the camera rays. 
}

\label{fig:synthetic_qualitative}
\end{center}
\end{figure*}
\topic{Cornea pose optimization}
Due to the small cornea size in the captured images, the cornea pose and normals estimate inevitably have some errors.
Training with the erroneous poses significantly affects the radiance field reconstruction's quality.
To alleviate the pose errors, we optimize the pose of each cornea independently. 
For each cornea, we optimize for a transformation matrix $T = \left[R, t \right] \in \text{SE(3)}$, where  $R \in \text{SO(3)}$  and $t \in \mathbb{R}^3$ denote the rotation and translation, respectively. 
We optimize the cornea poses during training similar to \cite{wang2021nerfmm,lin2021barf,meuleman2023localrf}.


\section{Experiments}
\begin{figure*}
\begin{center}
\centering

\mpage{0.99}{\includegraphics[width=\linewidth, trim=0 0 0 0, clip]{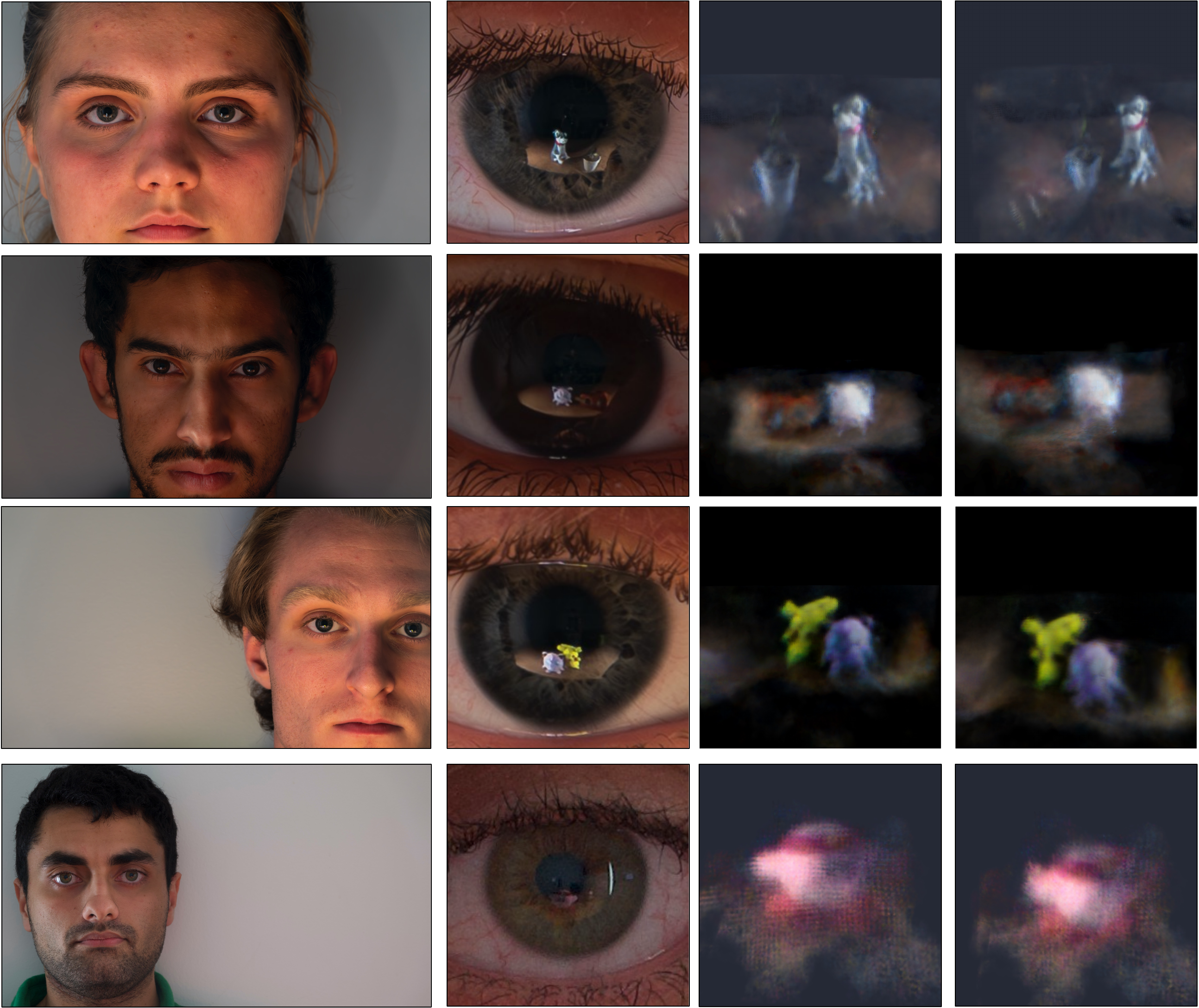}}\hfill
        \mpage{0.36}{Sample captured frame} \hfill 
    \mpage{0.19}{Eye crop} \hfill 
    \mpage{0.42}{$\underbrace{\hspace{\textwidth}}_{\substack{\vspace{-3.0mm}\\\colorbox{white}{~~Novel view rendering~~}}}$} \\
\captionof{figure}{
\textbf{Additional real results.} We show that our method works in a variety of capture conditions, like smaller objects as in the small plant on the top row, and varying eye colors. We show that we can also reconstruct the observed object with a significantly smaller eye observations like in the bottom example.}

\label{fig:additional_real}
\end{center}
\end{figure*}

\label{sec:exp}
\subsection{Synthetic data evaluation}
\begin{figure*}
\begin{center}
\centering

\mpage{1}{\includegraphics[width=\linewidth, trim=0 0 0 0, clip]{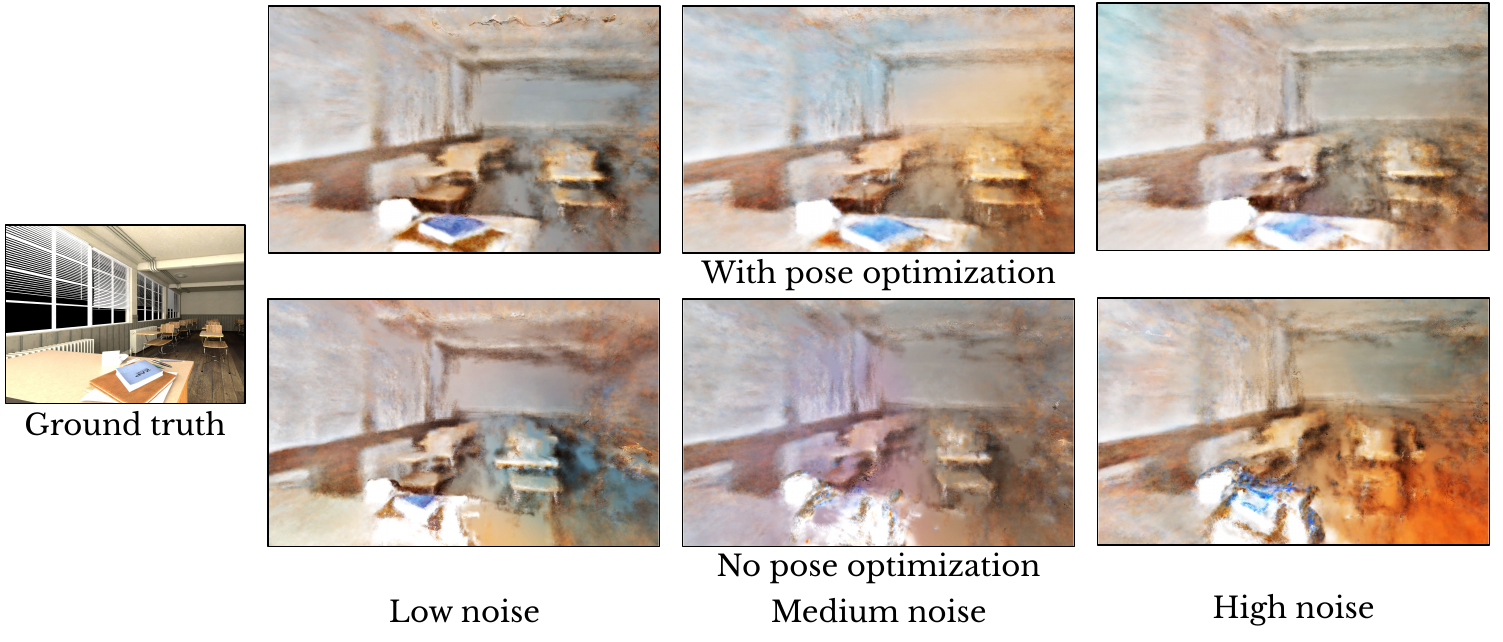}}\hfill
\captionof{figure}{
\textbf{Synthetic pose optimization ablation.} In simulation, the cornea pose optimization refines the noisy initial poses and results in clearer reconstruction.}

\label{fig:synthetic_pose_opt_ablation}
\end{center}
\end{figure*}

We generate synthetic data in Blender with eye models placed in the scene. In Figure \ref{fig:synthetic_qualitative} we show the scene we reconstruct using only the reflections from the eyes reflections.
Since we cannot estimate the cornea eye perfectly in real life, we evaluate the robustness of our cornea pose optimization to the noise in the estimated cornea radius.
To simulate the depth estimation errors we may encounter in real data, we corrupt the observed cornea $r_{img}$ radius for each image by scaling the estimated radius with varying noise levels.
In Figure \ref{fig:synthetic_pose_opt_ablation} we show how our method's performance varies for different noise levels.
Note that as the amount of noise increases, our reconstruction with pose optimization is robust in terms of the reconstructed geometry and colors when compared to the reconstruction without pose optimization. 
This demonstrates that pose optimization is essential for our method to work in realistic scenarios where the initial ellipse fitting in the image to the projected cornea is imperfect. 
Furthermore, we show quantitative comparisons of our method with and without texture decomposition in Table \ref{tab:texture_dec}. Our method performs better in terms of SSIM and LPIPS with texture decomposition than without. Notably, we do not compute PSNR because in our setting there is a drastic difference in lighting between the reflection and the scene itself.

\begin{table}[t]
\centering
\caption{
\textbf{Texture Decomposition Ablation.} We show that using a neural field to decompose the iris texture from the reflection improves reconstruction performance.
}
\resizebox{\linewidth}{!}{
\begin{tabular}{lcccc}
\toprule

\textbf{Scene} & \textbf{Method} & \textbf{SSIM $\uparrow$} & \textbf{LPIPS $\downarrow$} \\
\midrule
Classroom & w/o texture decomposition & 0.40  & 0.72 \\
& w/ texture decomposition & \textbf{0.42} & \textbf{0.62} \\
\midrule
Kitchen & w/o texture decomposition & 0.44 & 0.9 \\
& w/ texture decomposition & \textbf{0.48} & \textbf{0.82} \\
\bottomrule
\end{tabular}
}
\label{tab:texture_dec}
\end{table}

\subsection{Real-world experiments}
We describe capturing and processing real-world images and demonstrate the effectiveness of our method on real captures.
\begin{figure*}
\begin{center}
\centering

\includegraphics[width=0.95\linewidth, trim=0 0 0 0, clip]{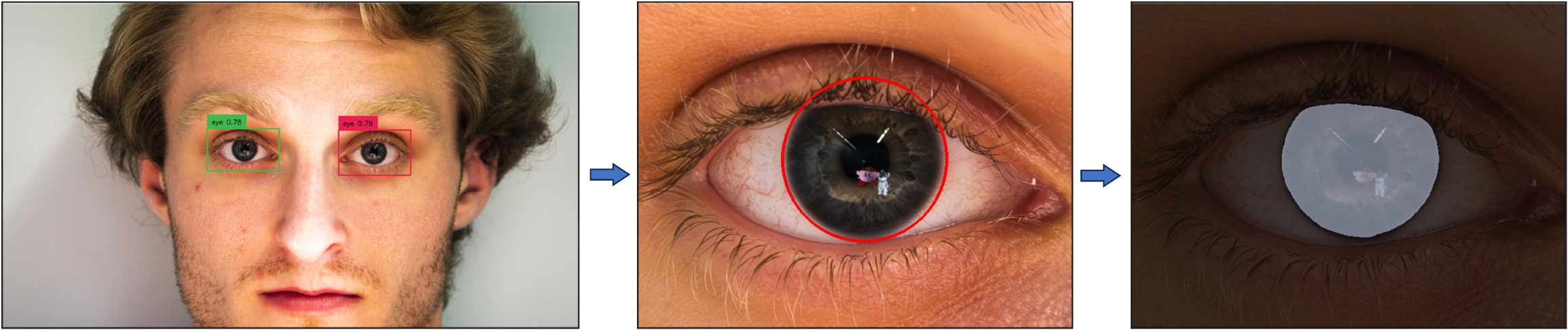}

    \mpage{0.37}{(a) Eye localization with GroundingDINO} \hfill 
    \mpage{0.31}{(b) Ellipse fitting with ELLSeg}
    \mpage{0.29}{(c) Iris segmentation with SAM}
\captionof{figure}{
\textbf{Data processing pipeline.} To compute the iris ellipse parameters, we first obtain eye bounding boxes using GroundingDINO \cite{liu2023grounding} and then conduct ellipse fitting using ELLSeg \cite{kothari2020ellseg}. Since we only want to use the visible regions of the cornea in our radiance field optimization, to handle occlusion, we generate a segmentation mask of the iris from the approximated cornea ellipse using Segment Anything \cite{kirillov2023segany}. 
}

\label{fig:data_processing}
\end{center}
\end{figure*}

\begin{figure}
\begin{center}
\centering

\frame{\includegraphics[width=\linewidth, trim=0 20cm 0 10cm, clip]{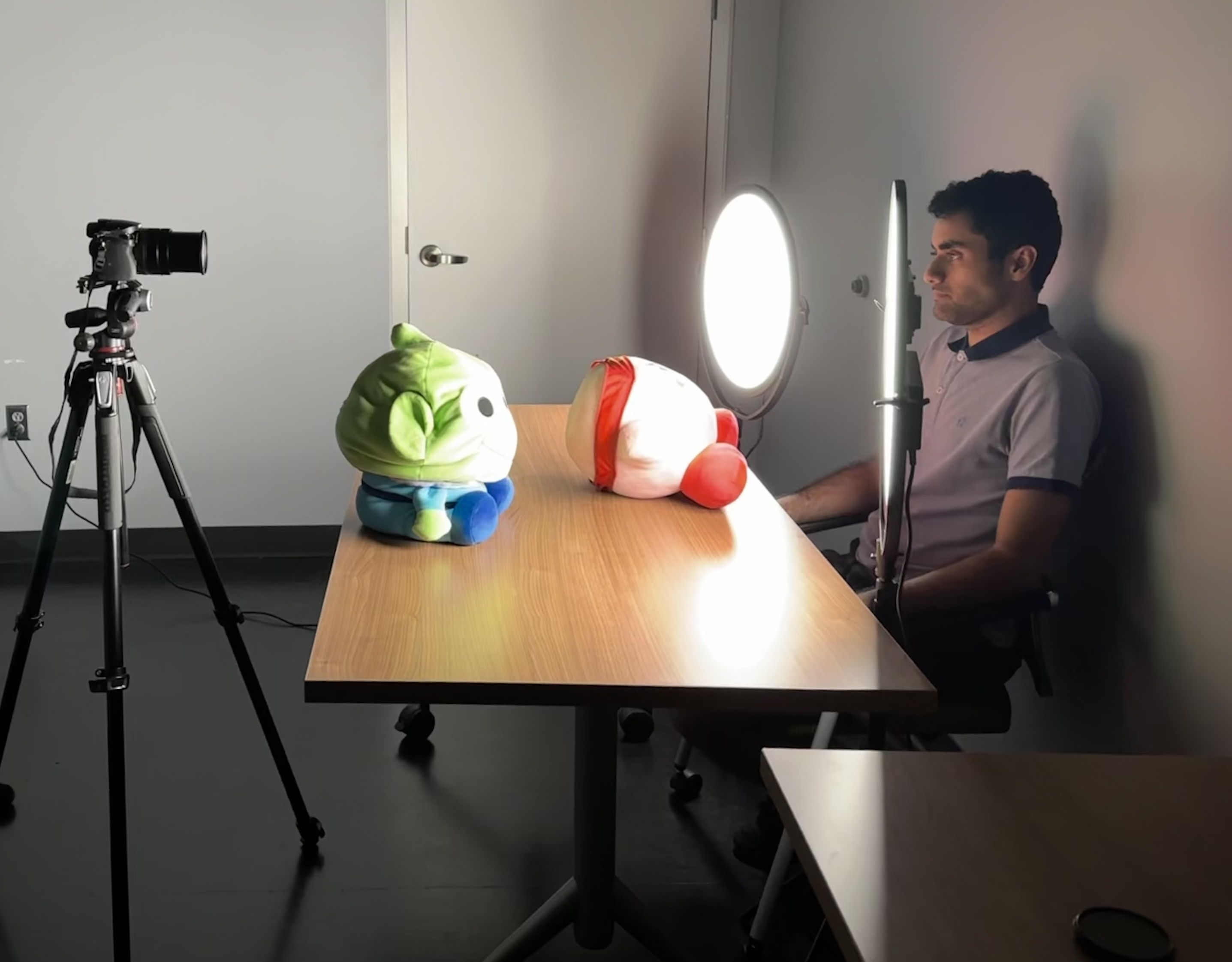}}

\captionof{figure}{
\textbf{Capture setup.} We illuminate the objects of interest with two area lights to ensure that sufficient amount of light is reflected off the eyes.
}

\label{fig:capture_setup}
\end{center}
\end{figure}
\topic{Image capture.} To maintain a realistic field of view, we capture images with a field of view that matches a standard portrait capture where the entire head is visible within the frame. We place area lights on the person's sides to illuminate the object of interest.
Figure \ref{fig:capture_setup} illustrates the capture setup. 
We ask the person to move within the camera's field of view and capture 5-15 frames per scene. 
We capture the images using a Sony RX IV camera and post-process the images using Adobe Lightroom to reduce the noise in the cornea's reflection. 
Since the captured images have a deep dynamic range due to the scene illumination, we use 16-bit images in all our experiments to avoid losing information from the observed reflections.
We vary the illumination brightness and the reflected object size for a comprehensive evaluation. 
On average, the cornea only covers around $0.1\%$ of each image, and the object of interest is reflected in a region of about 20x20 pixels and composited with the iris texture.

\subsubsection{Data processing}
\label{sec:processing}

We estimate the cornea's center and radius on images to get an initial estimate of the cornea's 3D location.
Once we have the radius, we can directly approximate the cornea's 3D location using the average depth from Eq. \ref{eq:depth} and the camera's focal length, and also compute its surface normals using Eq. \ref{eq:normal}. 
To automate the process, we locate the eyes bounding boxes using Grounding Dino \cite{liu2023grounding} and then use ELLSeg \cite{kothari2020ellseg} to perform ellipse fitting for the iris. 
While the corneas are typically occluded, we only need the unoccluded regions, so we obtain a segmentation mask for the iris using Segment Anything \cite{kirillov2023segany}. \\

\subsubsection{Results from real captures}
\begin{figure}
\begin{center}
\centering

\mpage{0.32}{\includegraphics[width=1.0\linewidth, trim=0 0 0 0, clip]{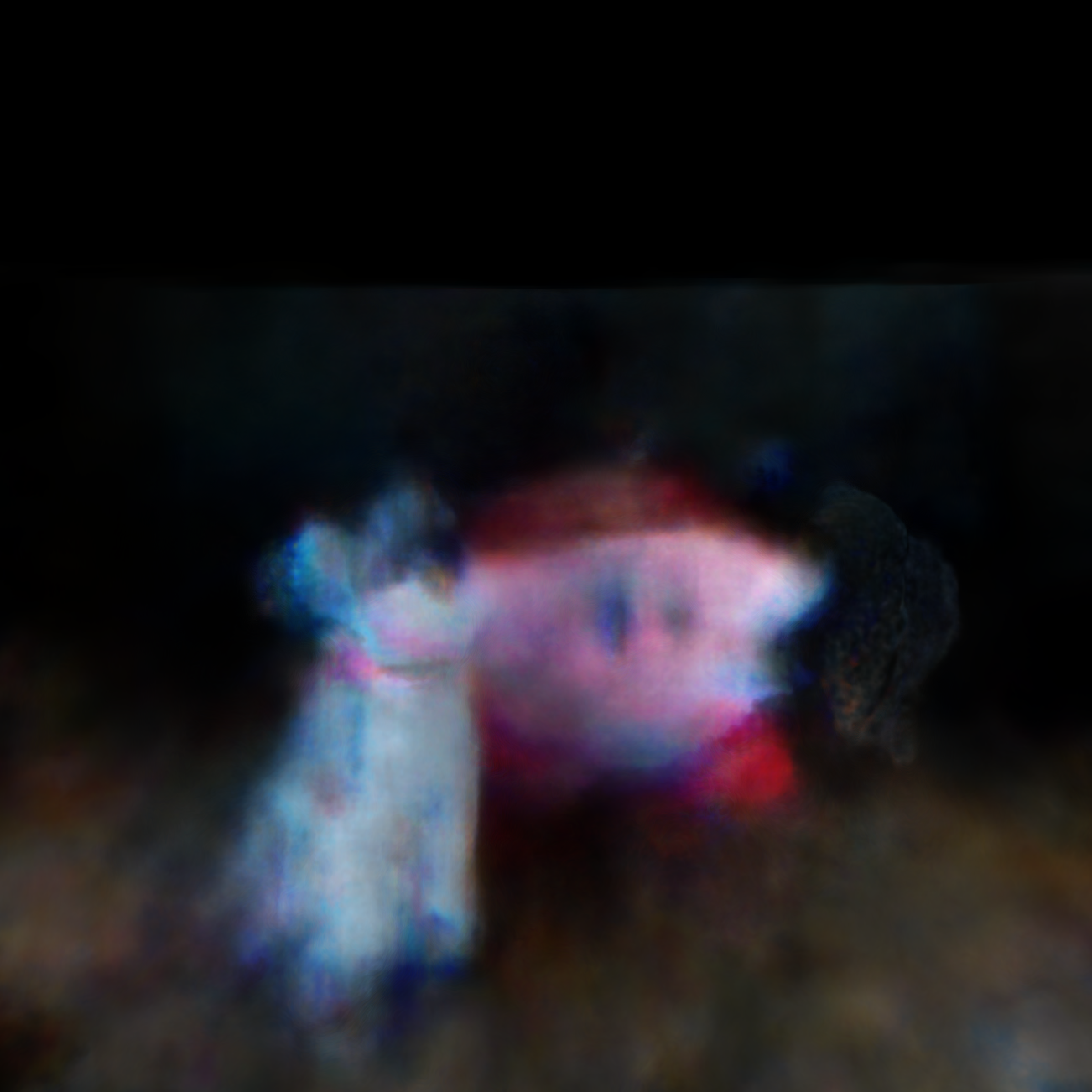}}\hfill
\mpage{0.32}{\includegraphics[width=1.0\linewidth, trim=0 0 0 0, clip]{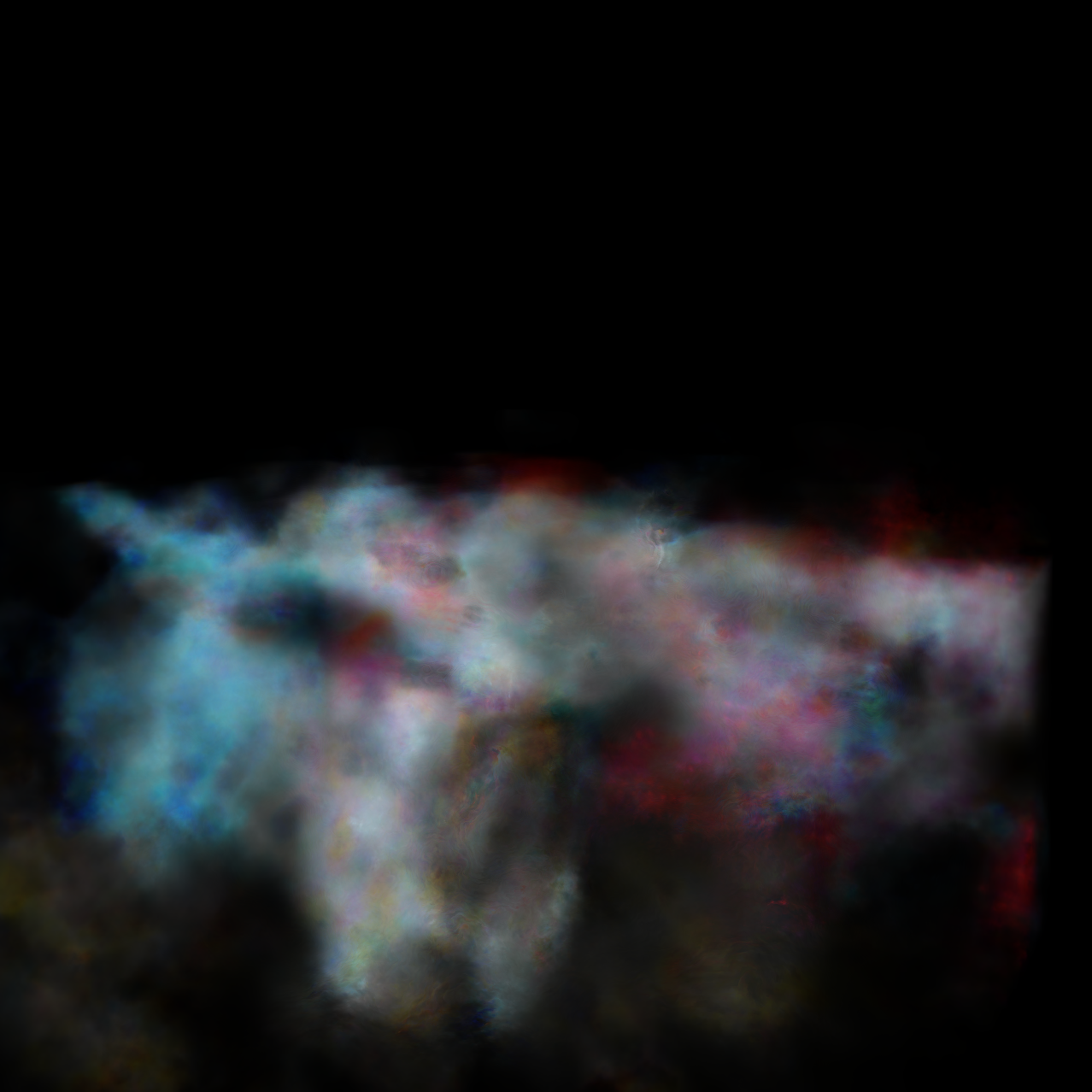}}\hfill
\mpage{0.32}{\includegraphics[width=1.0\linewidth, trim=0 0 0 0, clip]{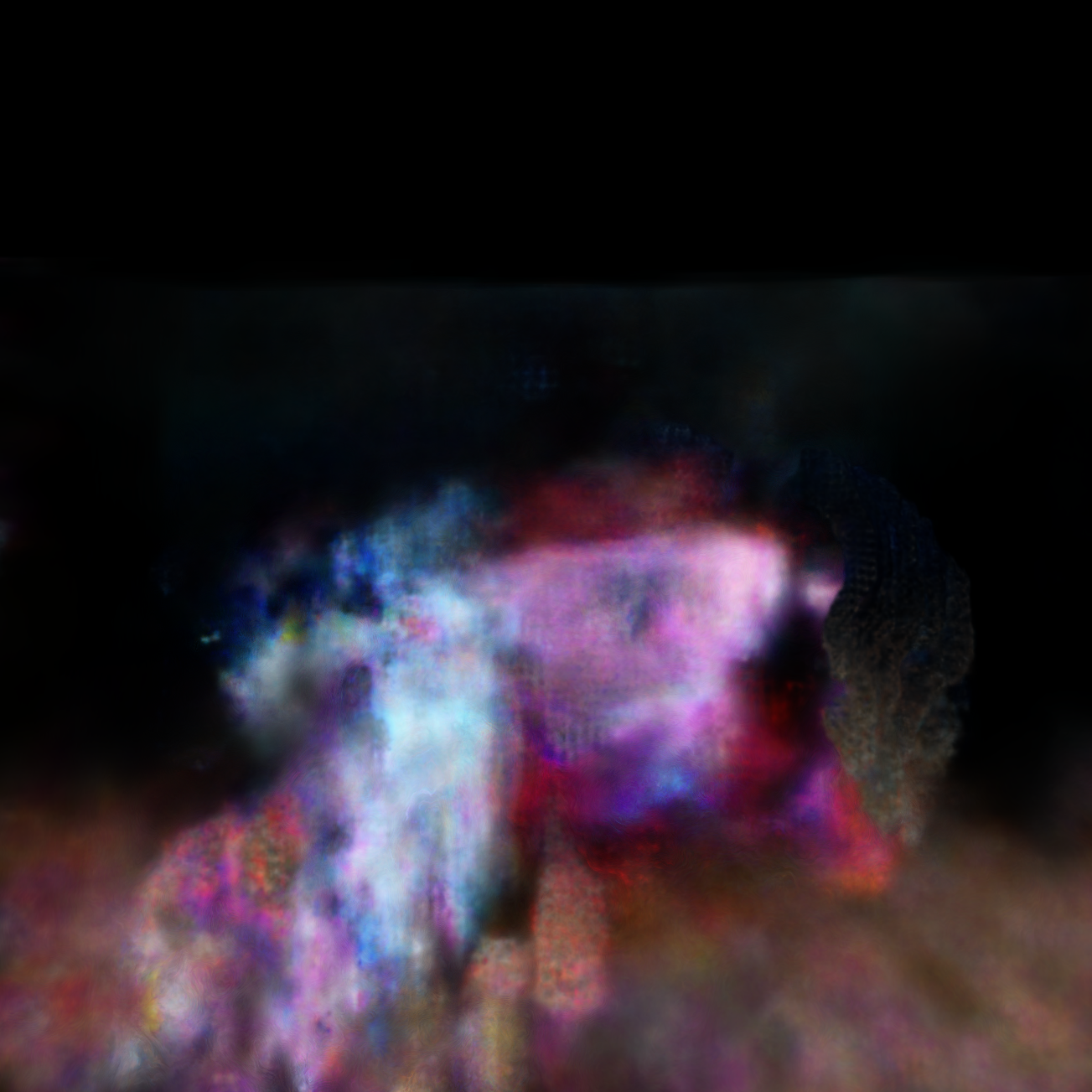}}\hfill
\mpage{0.32}{\includegraphics[width=1.0\linewidth, trim=0 0 0 0, clip]{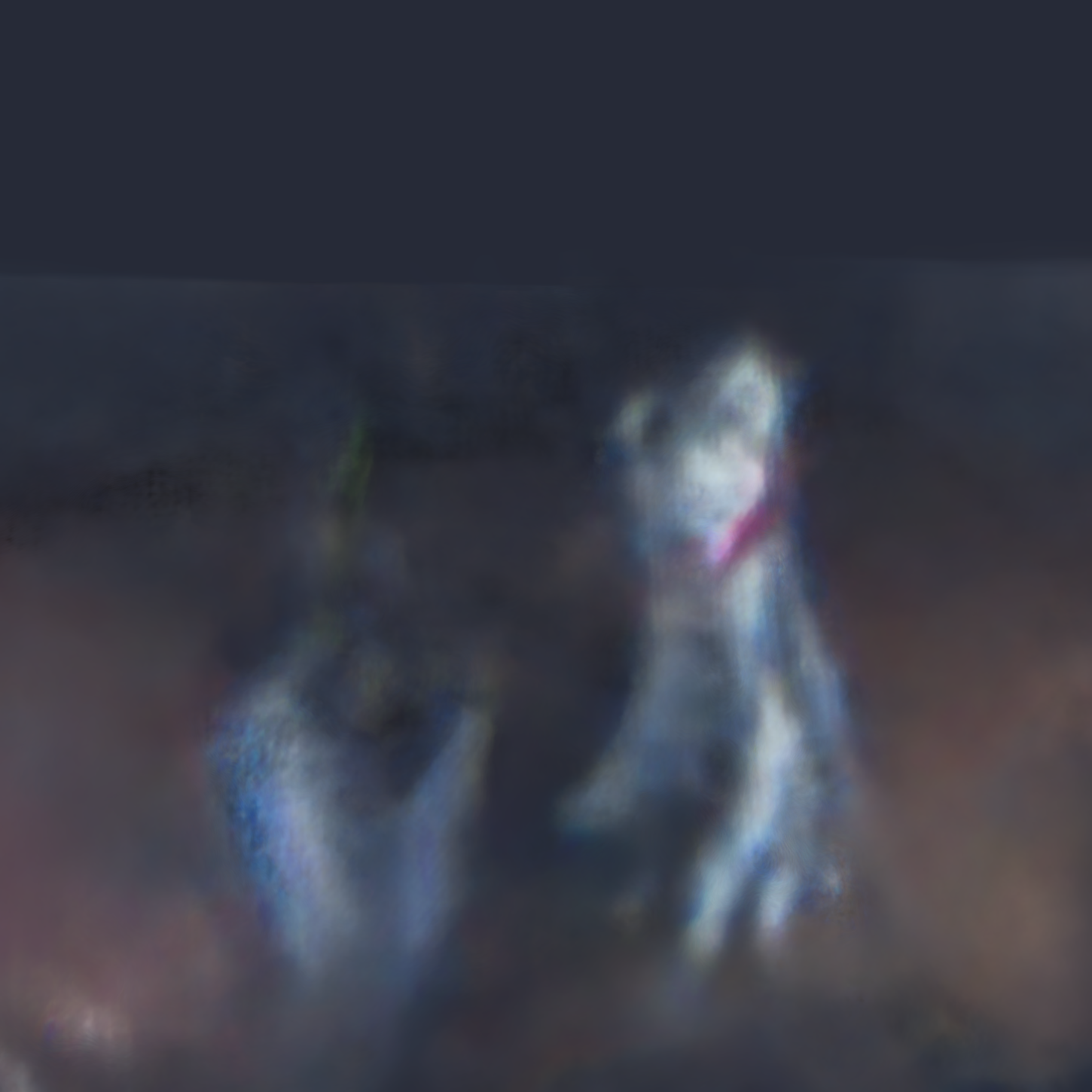}}\hfill
\mpage{0.32}{\includegraphics[width=1.0\linewidth, trim=0 0 0 0, clip]{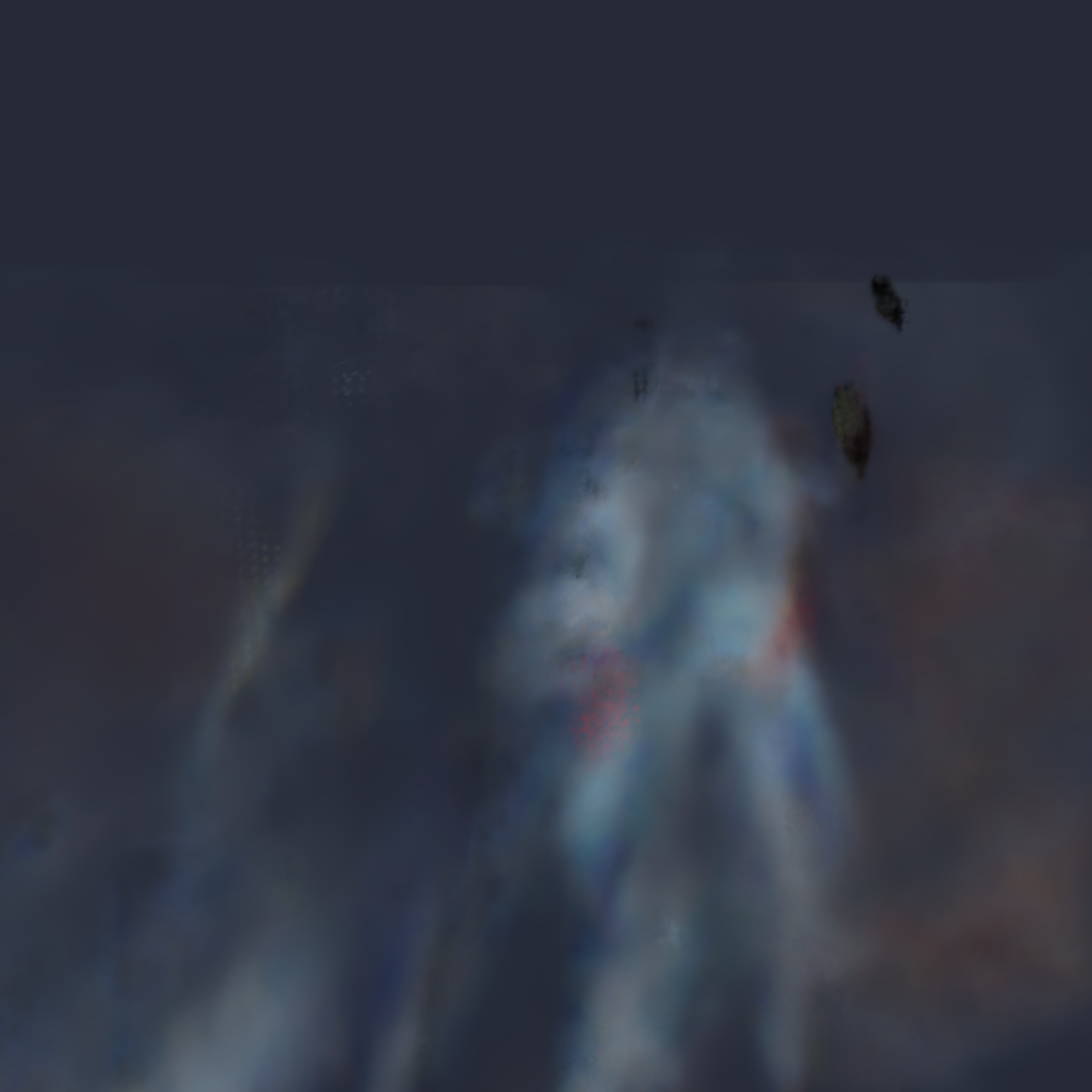}}\hfill
\mpage{0.32}{\includegraphics[width=1.0\linewidth, trim=0 0 0 0, clip]{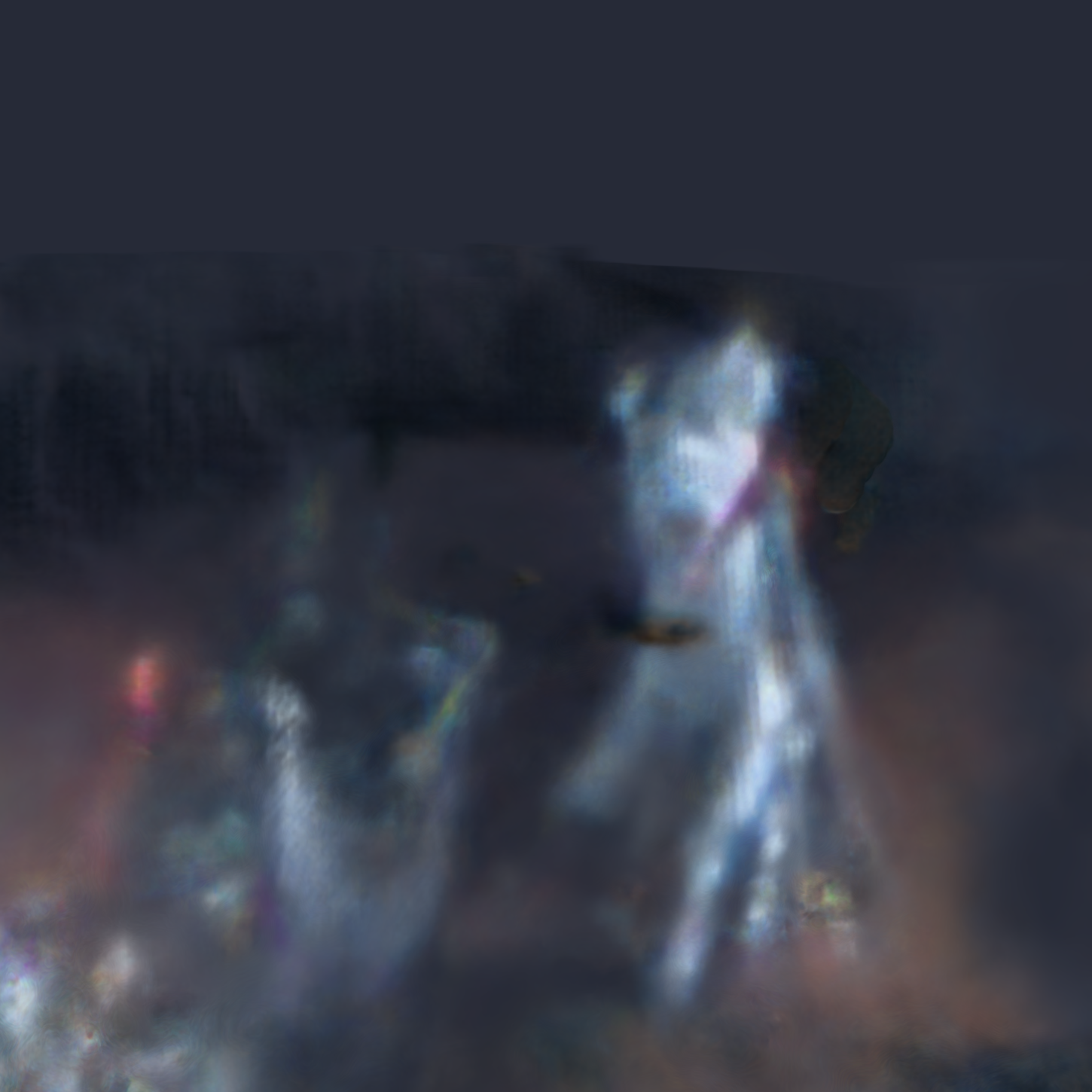}}\hfill
\mpage{0.31}{Full method} \hfill
\mpage{0.31}{No pose optimization} \hfill 
\mpage{0.31}{No texture decomposition}

\captionof{figure}{
\textbf{Ablating texture decomposition and cornea pose optimization.}
Top: not doing cornea pose estimation but still doing texture decomposition is not sufficient for 3D reconstruction at all. Bottom: not doing texture decomposition but still doing cornea pose estimation can recover some geometry and textures, but produces inferior visual quality. 
}
\label{fig:ablations}
\end{center}

\end{figure}

\begin{figure}[ht!]
\begin{center}
\centering

\mpage{1}{\includegraphics[width=\linewidth, trim=0 0 0 0, clip]{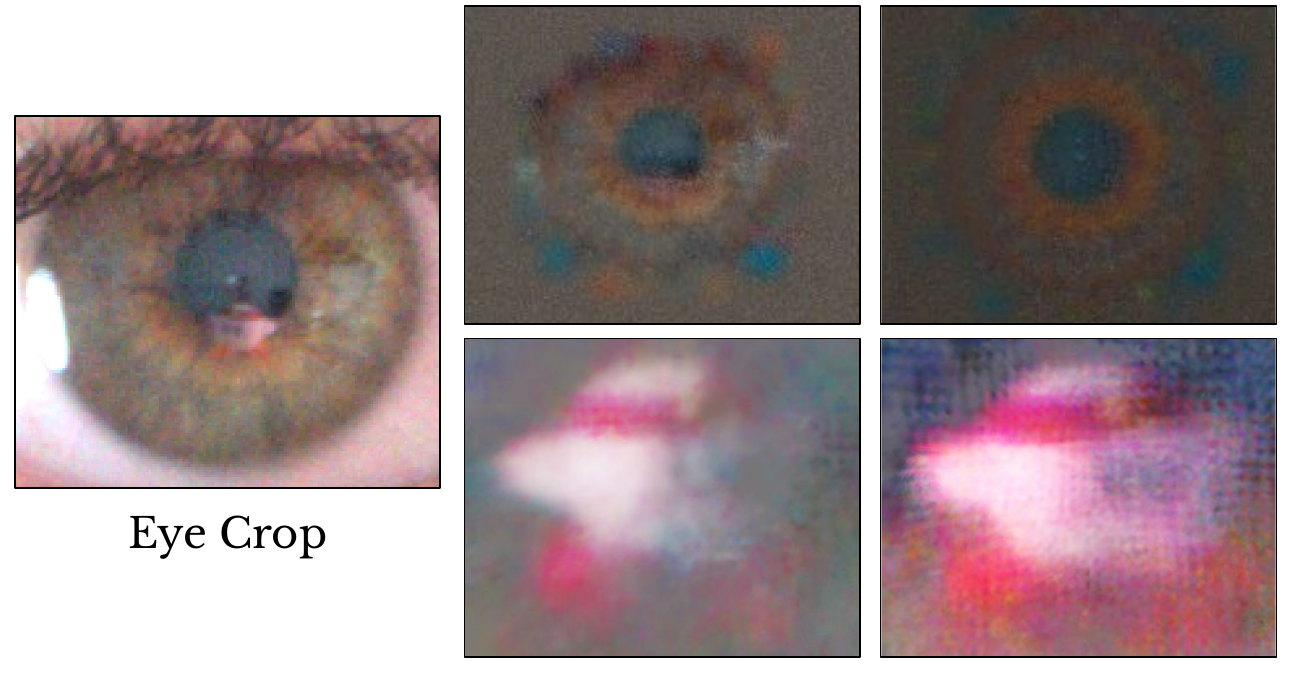}}\hfill
\mpage{0.3}{\text{ }} \hfill 
\mpage{0.3}{No $L_{radial}$} \hfill 
\mpage{0.3}{With $L_{radial}$}
\captionof{figure}{
\textbf{Ablating radial regularization.} 
Without radial regularization, the reconstructed iris texture contains parts of the scene with low disparity among observed views (e.g, here part of Kirby's body). Our radial loss alleviates this issue by penalizing the radial color variation of the iris texture. Notice that our synthesized view for the field trained with the radial prior has more detail than the view for the field trained without.} 
\label{fig:radial}
\end{center}

\end{figure}

Using our captured images, we show that our method enables the reconstruction of 3D scenes from real-world portrait captures, as shown in Figures \ref{fig:teaser} and \ref{fig:additional_real}, despite the cornea location and geometry estimate inaccuracies. 
In Figure \ref{fig:ablations}, by ablating the cornea pose optimization and texture decomposition from our method, we demonstrate that cornea pose optimization and texture decomposition are necessary for successful 3D scene reconstructions. 
The initial pose estimate of the corneas is noisy because the blurriness of the boundary of the cornea makes it challenging to be localized precisely in the image, as shown in Figure \ref{fig:radial}.
In Figure \ref{fig:ablations} we show the rendered radiance field with and without the learned texture decomposition. 
We notice significantly more floaters when not explicitly modeling the texture.
Furthermore, Figure \ref{fig:radial} demonstrates that the radial regularization improves the quality of our reconstruction because, without it, the texture decomposition will absorb parts of the scene with low disparity among observed views. 
We note that for some eye colors, like green and blue, the 3D reconstruction is more difficult because the iris texture is brighter. 
One such example of a green iris texture is given in Figure \ref{fig:radial}, and to handle these cases, we can increase the amount of radial regularization. 

\subsection{Limitations}
Our work demonstrates the feasibility of reconstructing the 3D world only from eye reflections. 
Two major limitations remain. 
First, our current real-world results are from a ``laboratory setup", such as a zoom-in capture of a person's face, area lights to illuminate the scene, and deliberate person's movement.
We believe more unconstrained settings remain challenging (e.g., video conferencing with natural head movement) due to lower sensor resolution, dynamic range, and motion blur.
Second, our assumptions on the iris texture (e.g., constant texture, radially constant colors) may be too simplistic so our approach may break down with large eye rotations.


\section{Conclusions}
\label{sec:conc}
By leveraging the subtle reflections of light off human eyes, we develop a method that can reconstruct the (non-line-of-sight) scene observed by a person using monocular image sequences captured at a fixed camera position. 
We demonstrate that naively training a radiance field on the observed reflections is insufficient due to several factors: 1) the inherent noise in cornea localization, 2) the complexity of iris textures, and 3) the low-resolution reflections captured in each image.
To address these challenges, we introduce cornea pose optimization and iris texture decomposition during training, aided by a radial texture regularization loss based on the nature of the human eye iris. 
We showcase the effectiveness of our approach to real-world data.
Unlike conventional methods of training a neural field that requires a moving camera, our method places the camera at a fixed viewpoint and relies solely on the user's motion.
With this work, we hope to inspire future explorations that leverage unexpected, accidental visual signals to reveal information about the world around us, broadening the horizons of 3D scene reconstruction.

{\small
\bibliographystyle{ieee_fullname}
\bibliography{egbib}
}

\end{document}



\title{
$\text{DC}^2$: Dual-Camera Defocus Control by Learning to Refocus \\Supplementary Material
}


\maketitle

 \newpage
 \setcounter{section}{0}
\renewcommand{\thesection}{\Alph{section}}

\clearpage
{\small
\bibliographystyle{ieee_fullname}
\bibliography{egbib}
}